# An Opinion Mining of Text in COVID-19 Issues along with Comparative Study in ML, BERT & RNN

Md. Mahadi Hasan Sany [1], Mumenunnesa Keya [1], Sharun Akter Khushbu[1], Akm Shahariar Azad Rabby [2], Abu Kaisar Mohammad Masum[1]

[1] Department of Computer Science and Engineering, Daffodil International University, Dhaka, Bangladesh
[2] The University of Alabama at Birmingham
{mahadi15-11173, mumenunnessa15-10100, sharun.cse, abu.cse}@diu.edu.bd
arabby@uab.edu

**Abstract.** The global world is crossing a pandemic situation where this is a catastrophic outbreak of Respiratory Syndrome recognized as COVID-19. This is a global threat all over the 212 countries that people every day meet with mighty situations. On the contrary, thousands of infected people live rich in mountains. Mental health is also affected by this worldwide coronavirus situation. Due to this situation online sources made a communicative place that common people shares their opinion in any agenda. Such as affected news related positive and negative, financial issues, country and family crisis, lack of import and export earning system etc. different kinds of circumstances are recent trendy news in anywhere. Thus, vast amounts of text are produced within moments therefore, in subcontinent areas the same as situation in other countries and peoples opinion of text and situation also same but the language is different. This article has proposed some specific inputs along with Bangla text comments from individual sources which can assure the goal of illustration that machine learning outcome capable of building an assistive system. Opinion mining assistive system can be impactful in all language preferences possible. To the best of our knowledge, the article predicted the Bangla input text on COVID-19 issues proposed ML algorithms and deep learning models analysis also check the future reachability with a comparative analysis. Comparative analysis states a report on text prediction accuracy is 91% along with ML algorithms and 79% along with Deep Learning Models.

## 1 Introduction

Sentiment analysis these days can be considered as one of the foremost famous inquiries about themes within the field of characteristic dialect preparation. Sentiment analysis is built on the utilize of natural language processing, biometrics, substance examination, and computational historical underpinnings to proficiently bring out, degree, recognize and dismember passionate conditions and personal data [1]. The employments of estimation examination are secured by a few curiously logical and commercial ranges, such as supposition mining, event detection, and recommender frameworks [2]. These days the social media stages such as Facebook, Twitter, WeChat, Instagram and YouTube, are an extraordinary field of data known as social information [3] besides that many people express their opinion on other platforms like different websites and survey administration software like Google-forms, etc. Other than that the rise in emphasis on AI strategies for literary analytics and NLP taken after the colossal increment in open reliance on social media like Twitter, Facebook, Instagram, blogging, and LinkedIn for information, rather than on the customary news organizations [4-6]. Coronavirus (COVID-19) which started to seem after 2019 in Wuhan, China has been one of the foremost examined and one of the foremost overthrowing infections all over the world. By the result of World Health Organization (WHO) until we are composing this think (2021/4/27) more than 14, 68, 41,882 people have contracted the malady and more than 31, 04,743 people have passed on in 223 countries over the world (WHO | World Health Organization). People’s lives are getting to be exceptionally dubious day by day as there's no treatable antibody. All through the history of mortality, social orders have been gone up against with crises of diverse sorts and of exceptionally distinctive natures, such as crises caused by the organization, send out of imperativeness resources, or emergencies caused by the impact of sicknesses and plagues, charitable clashes, and cash related crises, and among others [7]. We are in an exceedingly globalized world, where improvement flexibility, free travel between nations, and the headway and utilization of Data and Communication Developments (ICTs) are completely set up [8-9]. But within the show circumstance, the current crisis caused by COVID-19 is making a social mechanical situation that's exceptional for society [10]. In expansion, we must highlight the numerous countries that are locked down to urge a diminished rate of infection and death and many people are stuck in their houses. And these situations are responsible for many pyromaniacs. In this study, we’ve attempted to analyze the feelings or sentiments of Bangladeshi people in this emergency circumstance using several supervised machine learning (ML) algorithms and deep learning (DL) methods utilized to prepare our model. We scarcely found any work utilizing this kind of Bangla data-set on sentiment analysis. There are parcels of emotion analysis related work with Facebook, Reedit, Twitter, and many other social media stages data except for raw data collection from general people by Google Form. We’ve collected parts of Bangla feedback from 443 general people related to covid-19. In our input shape contain two sorts of questions. Within the to begin with address, there are six sorts of suppositions where there are three positive and three negatives, and within the moment address depressed or not depressed. People gave criticism based on those six suppositions and against those conclusions, they moreover gave the input that whether they are depressed or not. Authorities ought to know people internal conditions so that authority can persuade them. We have moreover compared all algorithms and models for a way better understanding. The abbreviations we have used in our work is given bellow-

**Table 1.** ML Acronyms.

| **Acronym** | **Definition** |
|---|---|
| MNB | Multinomial Naive Bayes |
| RFC | Random Forest Classifier |
| DTC | Decision Tree Classification |
| SVC | Support Vector Classifier |
| KNN | K-Nearest Neighbours |
| LR | Logistic Regression |
| RNN | Recurrent Neural Network |
| BERT | Bidirectional Encoder Representations from Transformers |

**Table 2.** Acronyms for Statistical Measurement.

| Acronym | Definition |
|---|---|
| CM | Confusion Matrix |
| MAE | Mean Absolute Error |
| MSE | Mean Square Error |
| RMSE | Root Mean Square Error |
| TP | True Positive |
| TN | True Negative |
| FP | False Positive |
| FN | False Negative |
| LL | Log Loss |

## 2 Literature Review

Sentiment analysis can be a computational bridge to people's internal assumptions or sentiments. Previously, numerous studies have been carried out on sentiment analysis, to name a few, prediction on political approaches through political articles, speeches [11], opinions of news entities on newspapers and blogs [12], stock exchange [13-14], cricket related comments from several social platforms using Bangla Text [15], etc. By utilizing data mining calculations, people's sentiment can be judged through social media. Recently, several groups of research scholars have devoted their effort to study and analyze information that is collected from numerous social media for different purposes using sentiment analysis. For this analysis many of them applied machine learning algorithms or deep learning models or both. Since the beginning of the widespread use of Covid-19, the work displayed by Aljameel et al. [16] pointed to conducting sentiment analysis on Saudi Arabia state's people of awareness on the Twitter-based data set. Nair et. al [17] they use BERT, VADER, and LR to find the sentiment of tweets whether it is positive or negative or neutral. In [18] Sabbir et. al they analyze on 1120 Bangla data using CNN and RNN to get output that is angry, depressed, or sad. Where CNN is able to get the most accurate and the model predicts that most people commented or expressed analytically. In [19] Shaika and Wasifa did sentiment analysis on Bangla Microblogs post. Rashedul et. al [33] approach an automated system where the system can categorise six individual emotion classes from Bangla text using supervised learning technique. In this pandemic situation, the Bangladesh government locked down properly to control the situation like other countries of the world. In this situation, people's mental situations are not good enough [20]. Most of the researchers collect their data from many social media and other sites, but we collect our data from general people by Google Forms where most of them are students from different universities at the age between 18-30 years age. Recently, Anjir et. al [34] used traditional ML algorithms and deep learning models such as RNN, RNNs, GRUs, BERT to analyze the sentiment of people about vaccination of Covid-19. They survey among the general people and data was collected by Google form. Galina et. al [35] also collect raw data by Google form to analyze the effect of Covid-19 pandemic on universities students learning.
We've utilized two classes to characterize the public's opinion which was collected by Google form. We've utilized a few machine learning algorithms and deep learning models for their better accuracy in our data set which is within the Bangla language.

# 3 Methodology

The outbreak of the Corona epidemic has had a huge impact around the world, especially in the population and among the elderly, adults, and middle-aged people. It has caused enough fear and anxiety in the world. So we did a survey of how people are spending their time in lockdown during this Covid-19 situation and what their mental state has been like while sitting at home. The methodology part is used to identify, select, process, and analyze information on any one subject in the study. In this part, we discuss in detail the research methodology such as - whether the qualitative or quantitative method of research has been used or why.

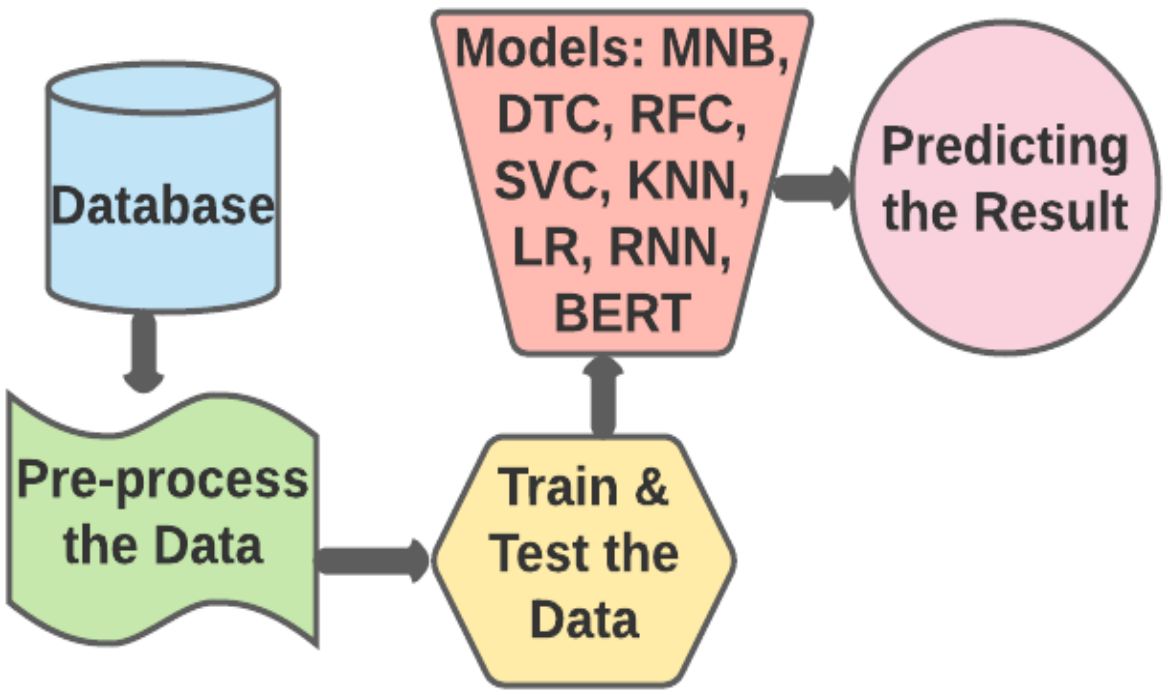

**Figure 1.** Proposed Model of Our Work. ([21]).

## 3.1. Data Collection

Data is the most important and vital part of Artificial Intelligence. The more we will give data the more machines can predict and analyse impeccably [22]. A machine's efficiency depends on the quality and the sum of information. In this proposed approach we've collected data from the general people using social media like Facebook and messenger chat groups by Google Form shown in figure-2. We were able to take 1085 responses based on 2 questions where the first question contains 6 opinions where 3 are positive and 3 are negative and based on these opinions, everyone responds in the second question whether they are depressed or not shown in Table-3. Finally, we've got two columns 'TypeOfOpinion', and 'Status' where 'TypeOfOpinion' refers to the opinions and 'Status' refers to the status of depression level. Among all of those responses of the moment address, we've found 550 under depressed, 535 not depressed.

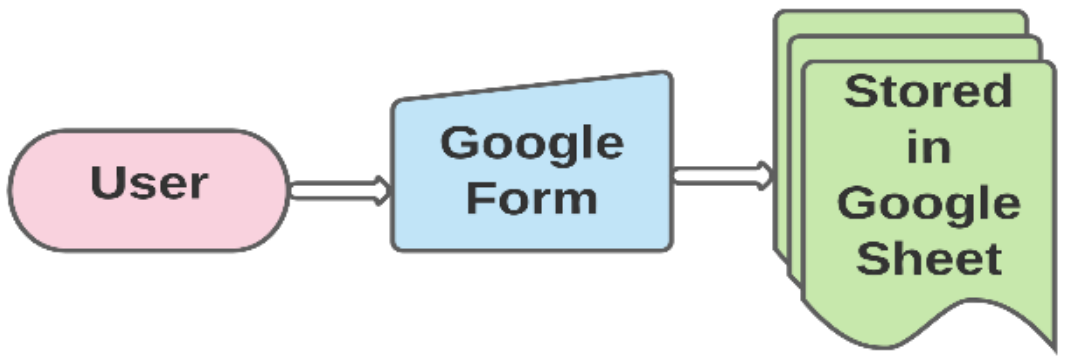

**Figure 2.** Google Form.

**Table 3.** Responses.

| **কভিড-১৯ এর সুবিধা এবং অসুবিধা (একাদিক অপশন সিলেক্ট করতে পারবেন) [Advantages and disadvantages of Covid-19 (you can select multiple option)] as TypeOfOpinion** | **আপনি কি মানসিকভাবে অবসাদগ্রস্ত ? [Are you mentally depressed?] as Status** |
|---|---|
| স্বাস্থ সচেতন হয়েছি, এই সময়ে সময়ে এক্সট্রা এক্টিভেটিস করার যথেষ্ট সময় পেয়েছি (Having become health conscious, I have had enough time to do extra activities from time to time) | Yes |
| স্বাস্থ সচেতন হয়েছি (I became health-conscious) | No |

### 3.2. Pre-processing

Data preprocessing is required tasks for cleaning the data and making it reasonable for a machine learning show which moreover increases the accuracy and productivity of a machine learning model because real-world information generally contains commotions, lost values, and possibly in an unusable arrangement which cannot be specifically utilized for machine learning models [23]. In preprocessing there are two parts one is preprocessing for ML algorithms and preprocessing for DL methods. Before applying ML algorithms, we did some preprocessing steps, at first, the “Status”-column attributes “yes” and “No” were considered to be “1” and “0” respectively. Then in the data set there are a lot of punctuation used which is needless like as “[(), $ %^&*+={}\[\]:\"|\'\~`<]”.

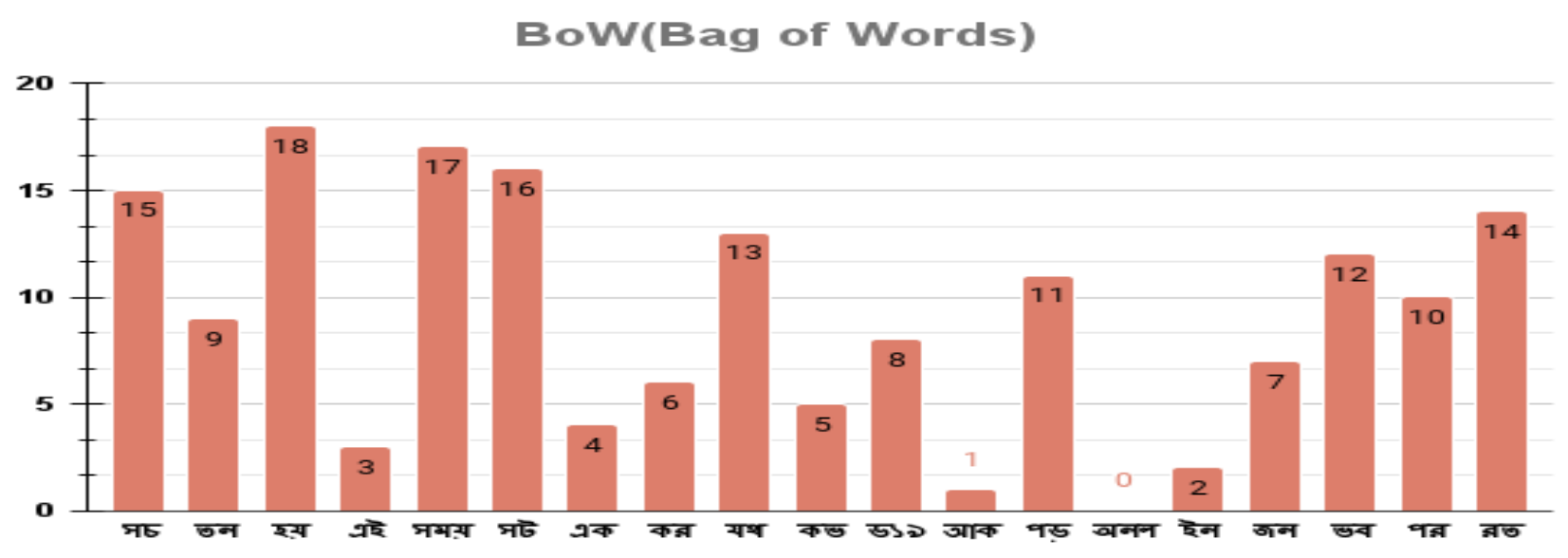

**Figure 3.** BoW

We remove this punctuation systematically by using libraries. After removing the punctuations, we remove the stop words. Stop word implies commonly used words like "[অতএব, অথচ, অথবা, অনুযায়ী, অনেক, অনেকে, অনেকেই, অন্তত, অন্য]" and so on. We've utilized the Bangla data set. Which had parcels of stop words. Those have been removed technically [24]. Then we utilize Bag of Words (BoW) in our data. BoW is utilized to change over a collection of content archives to a vector of term/token counts. It also empowers the pre-processing of content information earlier to produce the vector representation. This usefulness makes it a profoundly adaptable include representation module for content. After using BoW on my data set the segmented part for "স্বাস্থ সচেতন হয়েছি, এই সময়ে সময়ে এক্সট্রা এ একটিভেটিস করার যথেষ্ট সময় পেয়েছি, কভিড-১৯ এ আক্রান্ত হয়েছি, পড়ালেখা বিঘ্নিত হয়েছে" this sentence is like below figure.3. As like as ML here, we did some pre-processing step for feeding our RNN model, we implement tokenization. Tokenization fundamentally alludes to part up a bigger body of content into littler lines, words or indeed making words for a non-English dialect. The different tokenization capacities are in-built into the 'nltk' module itself and can be utilized in programs. For our data set, we use word tokenization to split the sentences into smaller pieces like: 'পড়', 'সময়', 'হয়' [25]. After that we labelencode the data set. LabelEncoder can encode a single variable. In this part, after tokenizing the whole data we encode the train and test part of the "Status" column by using LabelEncoder so that our model can do the further process for predicting accurately.Before feeding the model for train data BERT model expects input data in a specific format. In the BERT model "Tokenizer" does the preprocessing by itself. All the vectors ought to be of the same measure. Consequently, it should cushion the content to realize a common length. This length will be decided by the length of the longest sentence, which we'll need to calculate. Too, since we should concatenate the TypeOfOpinion and the Status column, the length ought to be chosen by the concatenated content column. Then tokenized the dataset that means splitting a sentence into a smaller as like: 'পড়', 'সময়', 'হয়, which is described above in the 'RNN' part. Including extraordinary tokens to check the starting ([CLS]) and separation/end of sentences ([SEP]). Then it creates token IDs. Pad the sentences to a common length. Create attention masks for the above PAD tokens.

### 3.3. Training Parameter

In table-4, the table is demonstrated for all information that is individually liable for the result which we actually got in result,

**Table 4.** Parameter Settings.

| Algorithms | RNN | | | BERT |
|---|---|---|---|---|
| **Train** 90% | **Train** 75% | **Beta-1** 0.9 | **Epoch** 15 | **Train** 80% |
| **Test** 10% | **Test** 25% | **Beta-2** 0.999 | **Verbose** 2 | **Test** 20% |
| **Random Size** 0 | **Learning Rate** 0.0005 | **Batch Size** 52 | **Validation Split** 0.15 | **Epoch** 3 |

Here, In the algorithm column, we took train size 90% and the test size was 10% cause our data set was small that's why we took the train size 90 so that the machine can learn more from the data set. And we took random state 0 cause if the value of the random state zero then the machine will take it as much as it needs. In the RNN column, it's noticeable that the train size is 75% and the test is 25%. For the RNN model, we split the data set 75:25 for getting better accuracy. Besides that, the learning rate, beta1, and beta2 are liable to get the accuracy that we demonstrated in table-7. In table-8 we can see that in epoch 13 we got 94% validation accuracy when verbose was 2 and validation split was 15%. The batch size was 52 which means per epoch 332(train size)/52(batch size) = 6.38 which means 6 steps will appear. Lastly in the BERT column train size is 80% and the test size is 20%. And the epoch size is 3, where 74 steps in per epoch.

### 3.3. RNN Model

Recurrent neural networks (RNN) is a sort of artificial neural network that includes extra weights to the network to make cycles within the network chart in an exertion to preserve an inside state. The promise of including state to neural networks is that they will be able to expressly learn and misuse setting in arrangement expectation issues, such as issues with an order or temporal component. In convolutional neural networks (CNN), all the inputs and outputs are autonomous of each other, but in cases like when it is required to predict the following word of a sentence, the last words are required and consequently, there's a have to be kept in mind the previous words. In this way, RNN came into existence, which solved this issue with the assistance of a Hidden Layer. The most vital feature of RNN is hidden, which recollects a few data almost an arrangement and RNN are a sort of Neural Network (NN) where the output from the last step is fed as input to the current step. RNN has a "memory" that recollects all data around what has been calculated. It employs the same parameters for each input because it performs the same errand on all the inputs or covered up layers to deliver the yield. This decreases the complexity of parameters, not at all like other neural networks.

### 3.4. BERT Model

BERT stands for Bidirectional Encoder Representations from Transformers. It is designed to pre-train significant bidirectional representations from the unlabeled substance by together conditioning on both left and right settings [32]. As a result, the pre-trained BERT show can be fine-tuned with a fair one extra output layer to make state-of-the-art models for a wide extent of NLP assignments. The BERT design builds on the best of Transformer. It has two variant right now, which are, BERT base & BERT large. Then it starts preprocessing with three combinations of embedding's which are Position Embedding, Segment Embedding and Token Embedding.

### 3.5. Statistical Analysis

Confusion Matrix: A confusion matrix may be a technique for summarizing the execution of a classification algorithm. Calculating a confusion matrix can give you a better idea of what a classification model is getting right and what sort of mistake it is making. The number of correct and erroneous predictions are summarized with tally values and broken down by each lesson. This can be the key to the confusion network. From confusion matrix we will able to get some values of precision, recall, and F1 score and equations are given bellow-

The equation of precision = $\frac{TruePositive}{True\ Positive+False\ Positive}$ [26]

The equation of recall = $\frac{True\ Positive}{True\ Positive+False\ Negative}$ [26]

The equation of F1 score = $2 * \frac{recision*recall}{precision+recall}$ [27].

## 4 Result and Discussion

This section will discuss all outcomes of both algorithms and the model part. Overall, we have worked on the humanoid condition of people during the Covid-19 period for which we have conducted a survey to find out about their human exhaustion. Through that survey, we found out if they are suffering from humanoid depression. And we are also aware of how they have benefited during the Coronation period.

***ML Algorithms-*** Considering the bar chart in Fig-4, it is seen that great results have been obtained by applying machine learning models (e.g. MNB, RFC, DTC, SVC, KN, and LR) for the survey. And it can be seen that among all the algorithms RFC gives 91% result which is more than other algorithms.

As we can see in Table-6, some steps of the Confusion Matrix are illustrated, between the states “0 and 1” of the dataset which indicates whether the person is mentally exhausted or not. Also shown is the status of the six algorithms, the values of precision, recall, F1 calculation, and support for “0 and 1”, which are seen differently for different algorithms. But in the case of “0”, it has the same support in each algorithm and has given another value for “1” but it is the same for all.

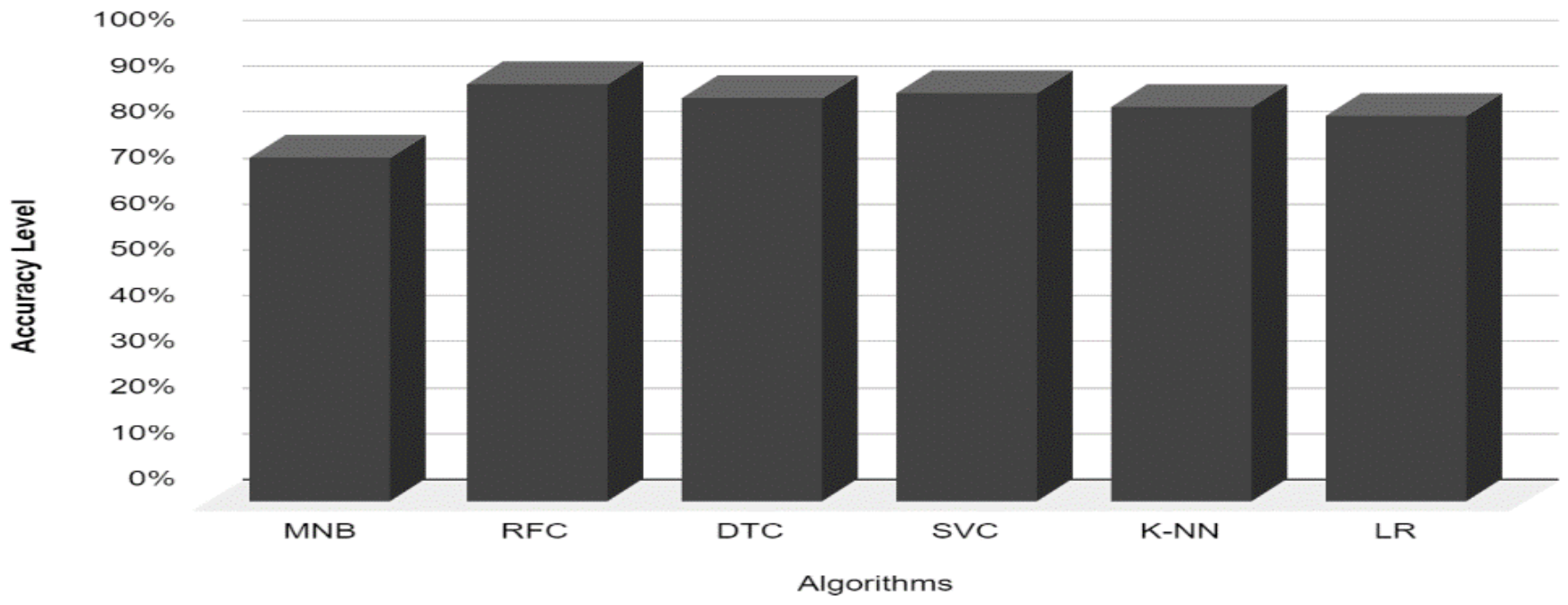

**Figure 4.** ML Algorithm Accuracy bar Chart**.**

As we can see in Table-7, the macro and weighted averages of the Confusion matrix are calculated for each algorithm for precision, recall, F1 score, and support. These values differ between the six algorithms, and in some cases, it returns the same value. As we can see, in the case of MNB, it has given the same values of Precision, F1 score, and support, but in the case of Recall, it has given different values, again in RFC, DTC, and SVC, it has given the same value to all. On the other hand, it has given different scores of precisions and F1 on KNN where it is different in terms of recall and support. All showed the same value except the F1 score on the LR. But for our proposed model SVC did better than K-NN wherein [29] K-NN did better than SVM.

Here, Table-8 demonstrates the False Positive (FP), False Negative (FN), Negative Prediction Value (NPV), False Discovery Rate (FDR), Mean Absolute Error, Mean Squared Error, Root Mean Squared Error, Root Mean Squared Error, and Log Loss. We have shown these scores in this table just for the six algorithms and in the case of the model, only Bert shows all the scores of the Confusion matrix as shown in Table-11, 12**.**

Sensitivity and Specificity of all algorithms which we use for our dataset. Sensitivity is used in Figure-5 because it can predict the outcome of variable decisions over a certain range. It creates specific sets for variables and can determine the analyst by which it influences the result through a change. It also uses sensitivity analysis to determine the effects of variables. The specification has been used only to increase the flexibility of the models. In the x-axis, we have represented the algorithms and placed the levels towards the y-axis.

Figures-(4-5) and Table-(6-8), show all the features when training size 0.9 and test size 0.1. The results that come after changing the standard of training and examination are summarized in Table-6 through table. When the values of training and testing were 0.9 and 0.1, the accuracy of the model was much higher which is very low in the case of Table-6. We wanted to change the quality of training and testing over and over again to see how accurate the models are. As a result, when we compare Figure-4**.** And Table-9, we can see how much the quality of training and testing varies in terms of accuracy.

**Table 6.** Result of Precision, Recall, F1-Score, Support.

| Algorithms | Status | Precision | Recall | F1-Score |
|---|---|---|---|---|
| **MNB** | 0 | 0.87 | 0.59 | 0.70 |
| | 1 | 0.70 | 0.91 | 0.79 |
| **RFC** | 0 | 0.91 | 0.91 | 0.91 |
| | 1 | 0.91 | 0.91 | 0.91 |
| **DTC** | 0 | 0.87 | 0.91 | 0.88 |
| | 1 | 0.91 | 0.87 | 0.89 |
| **SVC** | 0 | 0.90 | 086 | 0.89 |
| | 1 | 0.88 | 0.91 | 0.89 |
| **K-NN** | 0 | 0.86 | 0.86 | 0.86 |
| | 1 | 0.87 | 0.87 | 0.87 |
| **LR** | 0 | 0.94 | 0.73 | 0.82 |
| | 1 | 0.79 | 0.96 | 0.86 |

**Table 7.** Macro and Weighted Average for Precision, Recall, F1-Score, and Support.

| Algorithms | Macro & Weighted average | Precision | Recall | F1-Score |
|---|---|---|---|---|
| **MNB** | Macro avg. | 0.78 | 0.75 | 0.75 |
| | Weighted avg. | 0.78 | 0.76 | 0.75 |
| **RFC** | Macro avg. | 0.91 | 0.91 | 0.91 |
| | Weighted avg. | 0.91 | 0.91 | 0.91 |
| **DTC** | Macro avg. | 0.89 | 0.89 | 0.89 |
| | Weighted avg. | 0.89 | 0.89 | 0.89 |
| **SVC** | Macro avg. | 0.89 | 089 | 0.89 |
| | Weighted avg. | 0.89 | 0.89 | 0.89 |
| **K-NN** | Macro avg. | 0.87 | 0.87 | 0.87 |
| | Weighted avg. | 0.86 | 0.87 | 0.86 |
| **LR** | Macro avg. | 0.86 | 0.84 | 0.84 |
| | Weighted avg. | 0.86 | 0.84 | 0.85 |

**Table 8.** Confusion Matrix for All Algorithm.

| Algorithm | FP | FN | NPV | FDR | MAE | MSE | RMSE |
|---|---|---|---|---|---|---|---|
| **MNB** | 0.3 | 0.13 | 0.91 | 0.41 | 0.24 | 0.25 | 0.49 |
| **RFC** | 0.08 | 0.09 | 0.91 | 0.09 | 0.08 | 0.08 | 0.29 |
| **DTC** | 0.09 | 0.13 | 0.86 | 0.09 | 0.11 | 0.11 | 0.33 |
| **SVC** | 0.12 | 0.09 | 0.91 | 0.13 | 0.11 | 0.11 | 0.39 |
| **K-NN** | 0.13 | 0.14 | 0.86 | 0.14 | 0.13 | 0.13 | 0.36 |
| **LR** | 0.21 | 0.05 | 0.95 | 0.27 | 0.15 | 0.15 | 0.39 |

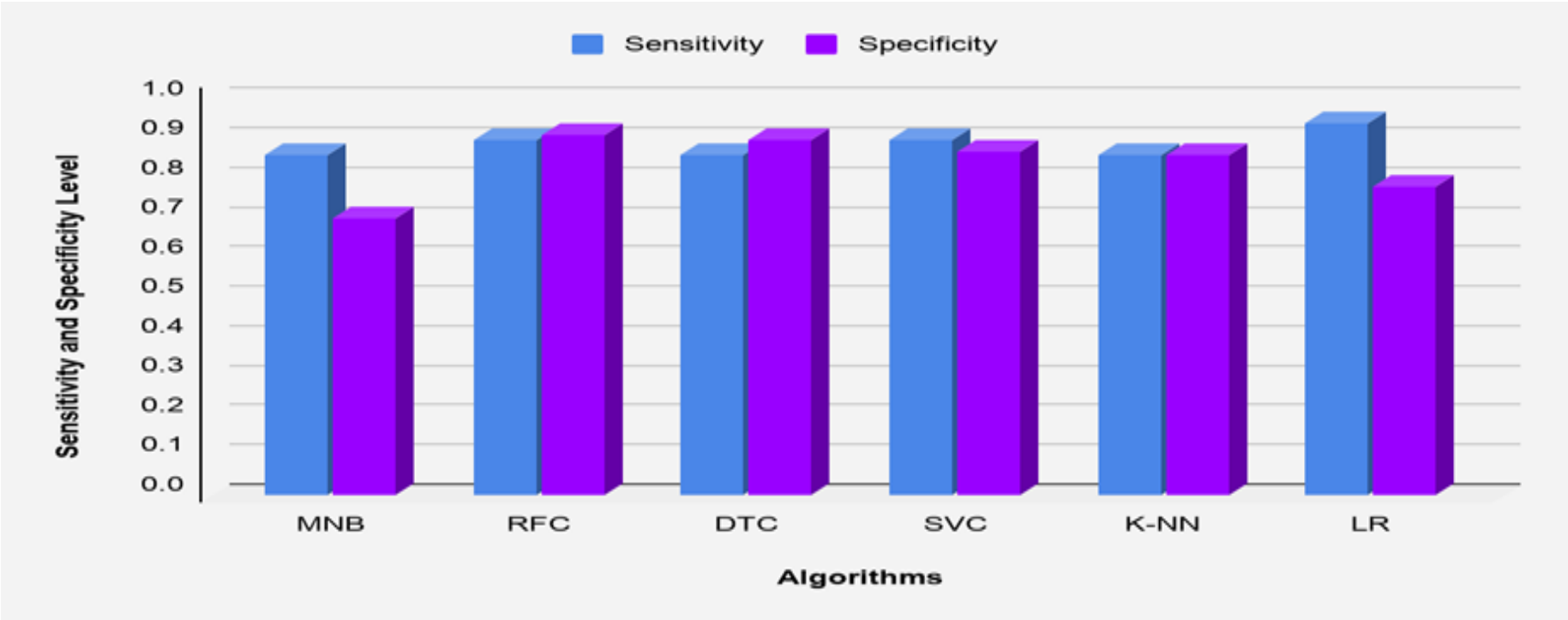

**Figure 5.** Sensitivity and Specificity bar Chart.

**Table 9.** Comparative Accuracy for ML Algorithms.

| **Train/Test Size** | **MNB** | **RFC** | **DTC** | **SVC** | **K-NN** | **LR** |
|---|---|---|---|---|---|---|
| Train = 0.50<br>Test = 0.50 | 72.97 | 68.01 | 68.47 | 67.12 | 69.82 | 68.92 |
| Train = 0.60<br>Test = 0.40 | 67.42 | 69.01 | 73.03 | 66.85 | 74.16 | 67.98 |
| Train = 0.70<br>Test = 0.30 | 74.44 | 74.44 | 73.68 | 67.70 | 75.19 | 67.70 |
| Train =0 .75<br>Test =0 .25 | 75.68 | 76.58 | 73.87 | 65.77 | 77.48 | 64.87 |
| Train = 0.80<br>Test = 0.20 | 74.15 | 79.78 | 78.65 | 67.41 | 69.66 | 67.41 |

***DL Models-*** This part describes all of the experimental results for our RNN and BERT models. After applying both models with the data set accuracy and loss result of train and test are shown in Table-10. Table-10 is the comparative study with the six algorithms of Figure-4. Table-10 assumes the RNN & BERT model train, test accuracy, loss, and all accuracy. We can see that the ML models give the best accuracy rather than the DL models. Between the two models BERT performs the best, it's both train test accuracy is higher than RNN and also the loss is lower than the other model.

**Table 10.** Accuracy and Loss of RNN and BERT model.

| **Models** | **Train** | | **Test** | | **All Accuracy** |
|---|---|---|---|---|---|
| **RNN** | 0.72 | 0.54 | 0.71 | 0.55 | 72% |
| **BERT** | 0.73 | 0.53 | 0.53 | 0.51 | 79% |

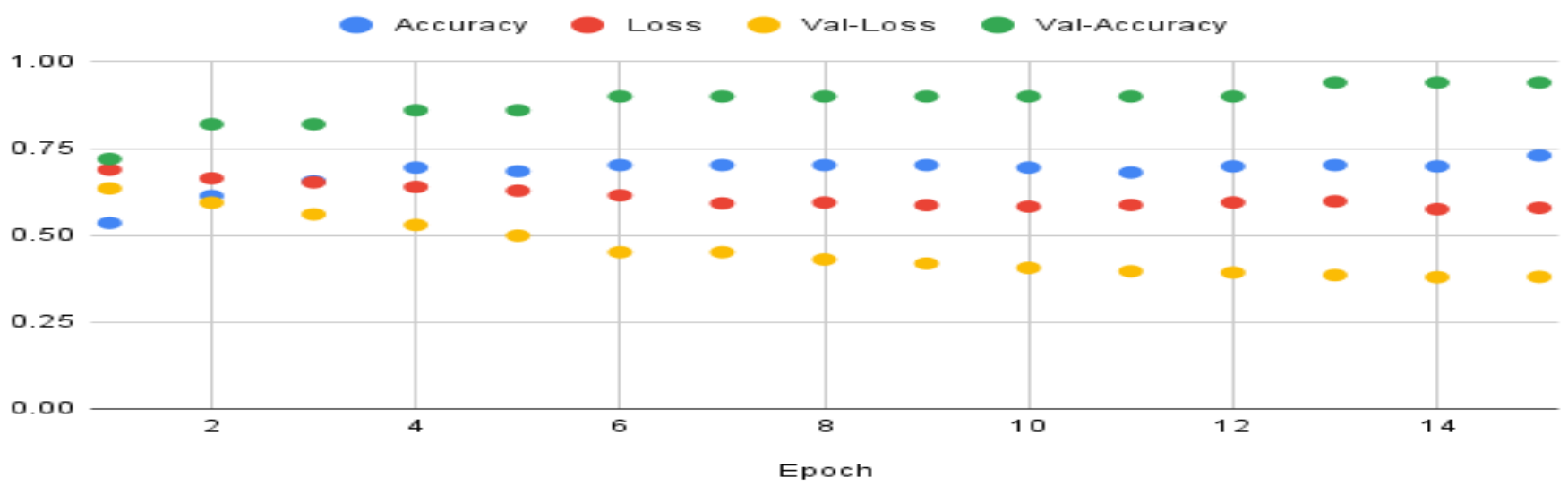

**Figure 6.** RNN Model History.

Model History of RNN is demonstrated in Figure. 6., 15 epochs have been used to run the model because after that it gives static results. That's why it has been kept in epoch 15. And it can be seen from Figure. 6 that over time the "accuracy" of the model, the "loss", the "validation accuracy" and the "validation loss", are getting more and less. In the case of RNN, we have got a maximum accuracy of 72% with a validity of 94% which is for very good modeling. From Figure. 6., it's noticeable that in the "Loss" column reducing rate is slower than the "Val-Loss" column. On the other hand, the increasing rate of "Val-Accuracy" is higher than "Accuracy". Figure. 7 represents the ratio of training and validation accuracy and loss graph. From the graph, we can see that the accuracy is less than validation accuracy and also loss is less than the model accuracy.

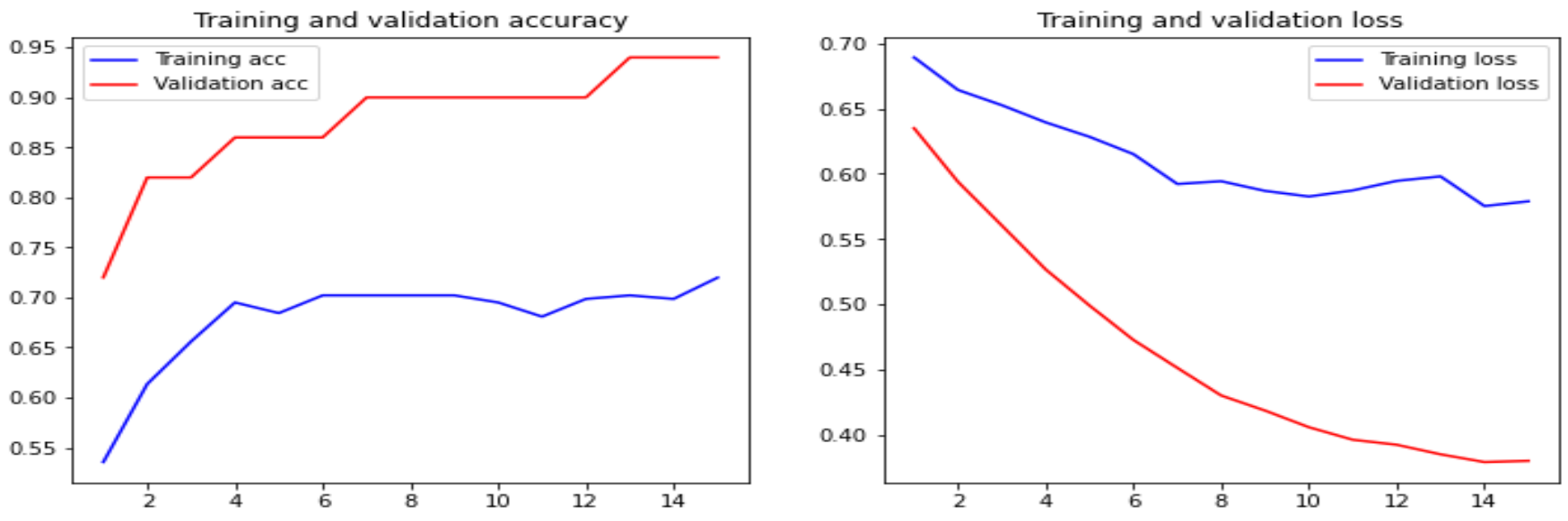

**Figure 7.** Graphical Representation of RNN

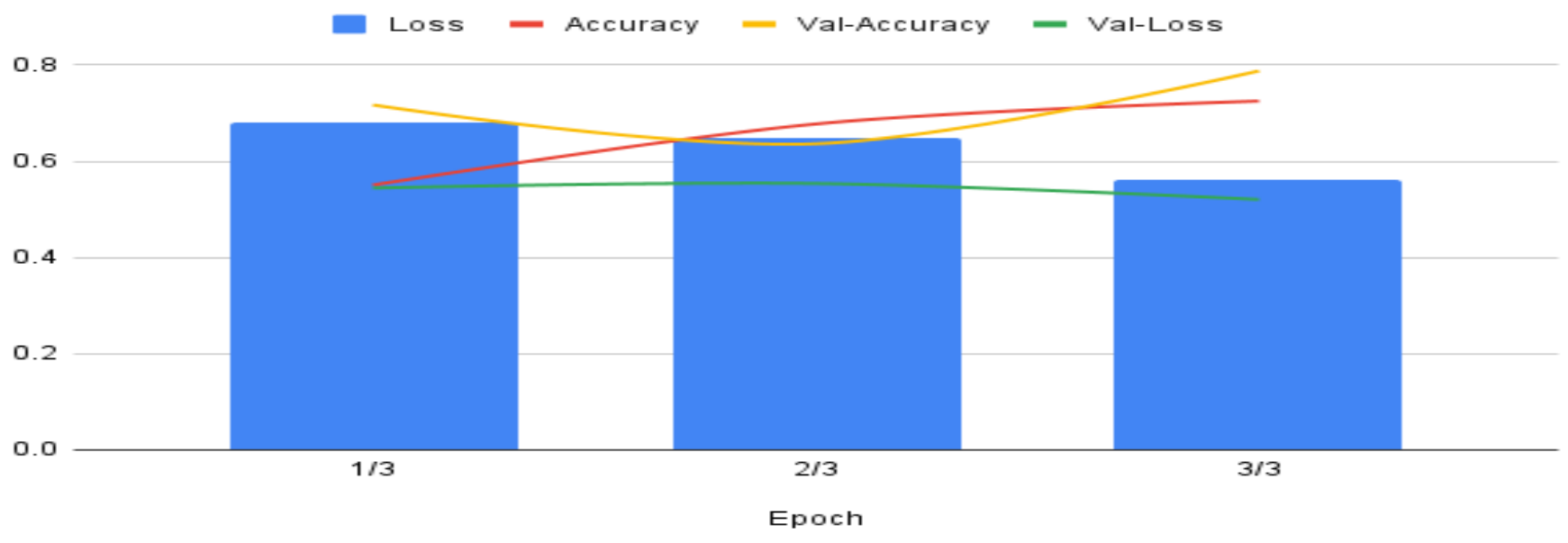

**Figure 8.** BERT model History.

**Table 11.** Precision, Recall, F1-Score, and Support.

| Model | Status | Precision | Recall | F1-Score |
|---|---|---|---|---|
| BERT | 0 | 0.72 | 0.86 | 0.78 |
| | 1 | 0.87 | 0.73 | 0.79 |

**Table 12.** Precision, Recall, F1-Score.

| Model | Macro & Weighted average | Precision | Recall | F1-Score |
|---|---|---|---|---|
| BERT | Macro average | 0.79 | 0.80 | 0.79 |
| | Weighted average | 0.80 | 0.79 | 0.79 |

The scenario is like column "Loss". In columns "Accuracy" and "Val-Accuracy", there the 2nd epoch value is lower than the 1st epoch and the 3rd epoch is higher than both the 1st and 2nd columns, and the whole scenario is different from the table-8 which is for the RNN model.

Here, table-11 representing the Precision, Recall, F1-Score, and Support of "Status". The resultant value describes that for "1" the precision is higher than "0" and recall is the opposite of precision. F1-Score and Support result difference is almost the same. Comparing Table-6 BERT gives almost a lower result for Classification report because BERT always works on big data but our data source is too short to run the model but still, it performs better in the small dataset.

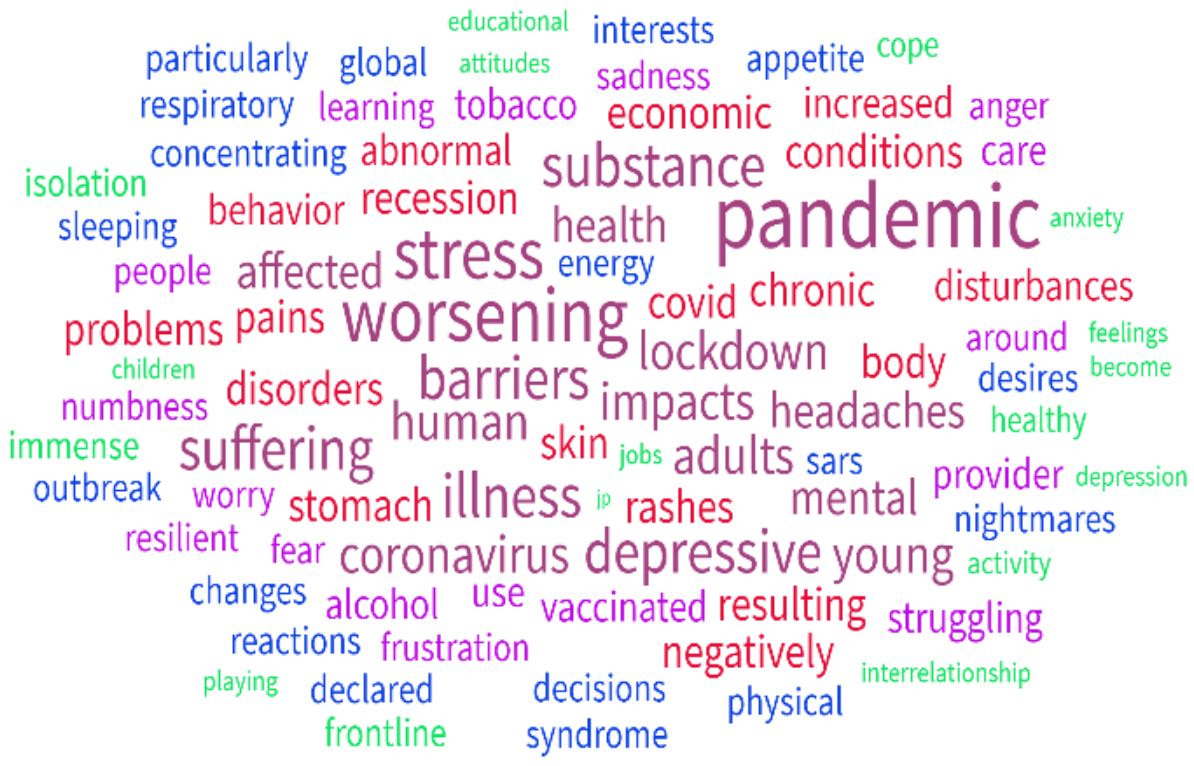

**Figure 9.** Word Cloud of Mental Health during Covid-19 Pandemic.

Support of macro and Weighted average from the BERT model. As in table-12 the weighted average is higher than the Macro average for precision and Recall is the opposite of precision and the F1-Score and Support result difference is the same Macro &

Weighted average. Table-12 & Table-7 comparatively almost equal and for short data, it's a big thing.
A word cloud is a grouping of words that are represented in various sizes. The larger and bolder the phrase, the more often it appears in a text and the more significant it is. In figure-9, the word cloud is a mental health condition during the Covid-19 situation where the most significant words of this topic are mentioned and the words collect from our dataset showed in the word cloud.

## 5 Conclusion and future work

From this experiment, the proposed model can predict the sentiment of a Bangla sentence whether it’s depressed or not. This model is proposed on only 443 data which was collected by Google form. Here we applied several supervised machine learning classification algorithms and also deep learning models which were RNN and BERT. Before applying these algorithms and models some preprocessing techniques and some heterogeneous features to feed our algorithms and models. Then we got 71% accuracy on deep learning-based model RNN and 79% on BERT besides that on algorithm portion RFC achieved 91%, SVC 89%, DTC 88%, K-NN 86%, LR 84%, and MNB 75% which is the lowest among all algorithms and RFC is the highest among models and algorithms. This experiment will conclude that by the victimization of this different kind of machine learning classification algorithm we will get better results, and conjointly using the deep learning models achieved very high accuracy even in an exceedingly touch of training data. We have some circumscription that the whole experiment was did on 443 data. And the data set was based on 6 opinions and there were 2 categories against the opinions whether it is depressed or not depressed. In the future, we'll experiment with totally diverse preprocessing strategies, heterogeneous features, supervised and unsupervised calculations for creating an extra exact framework and also work with social media related data. For stirring giant data; we will use projected heterogeneous options and deep learning options like Word2Vec and Word Embedding to apply deep learning algorithms like Long short-term memory (LSTM) [30] and Convolutional neural networks (CNN) [31] to induce outstanding results.